\documentclass[10pt,twocolumn,letterpaper]{article}

\usepackage{cvpr}
\usepackage{times}
\usepackage{epsfig}
\usepackage{graphicx}
\usepackage{amsmath}
\usepackage{amssymb}
\usepackage{amsfonts}
\usepackage{caption}
\usepackage{amsmath,amssymb,amsthm,mathrsfs,amsfonts,dsfont} 

\usepackage[breaklinks=true,bookmarks=false,colorlinks=true]{hyperref}

\cvprfinalcopy 

\def\cvprPaperID{369} 

\ifcvprfinal\fi
\begin{document}


\title{Total Capture: A 3D Deformation Model for Tracking Faces, Hands, and Bodies\thanks{Website:~\scriptsize{\url{http://www.cs.cmu.edu/\~hanbyulj/totalcapture}}}}



\author{Hanbyul Joo\hspace{0.3in} Tomas Simon\hspace{0.3in} Yaser Sheikh\\
Carnegie Mellon University\\
{\tt\small \{hanbyulj,tsimon,yaser\}@cs.cmu.edu}
}
\maketitle

\begin{abstract}
We present a unified deformation model for the markerless capture of multiple scales of human movement, including facial expressions, body motion, and hand gestures. An initial model is generated by locally stitching together models of the individual parts of the human body, which we refer to as the ``Frankenstein'' model. This model enables the full expression of part movements, including face and hands by a single seamless model. Using a large-scale capture of people wearing everyday clothes, we optimize the Frankenstein model to create ``Adam". Adam is a calibrated model that shares the same skeleton hierarchy as the initial model but can express hair and clothing geometry, making it directly usable for fitting people as they normally appear in everyday life. Finally, we demonstrate the use of these models for total motion tracking, simultaneously capturing the large-scale body movements and the subtle face and hand motion of a social group of people.
\end{abstract}

\section{Introduction}
Social communication is a key function of human motion \cite{Birdwhistell70}. We communicate tremendous amounts of information with the subtlest movements. Between a group of interacting individuals, gestures such as a gentle shrug of the shoulders, a quick turn of the head, or an uneasy shifting of weight from foot to foot, all transmit critical information about the attention, emotion, and intention to observers. Notably, these social signals are usually transmitted by the organized motion of the whole body: with facial expressions, hand gestures, and body posture. These rich signals layer upon goal-directed activity in constructing the behavior of humans, and are therefore crucial for the machine perception of human activity. 

However, there are no existing systems that can track, without markers, the human body, face, and hands simultaneously. Current markerless motion capture systems focus at a particular scale or on a particular part. Each area has its own preferred capture configuration: (1) torso and limb motions are captured in a sufficiently large working volume where people can freely move~\cite{deAguiar-2008, Gall-09, Stoll-11, Elhayek-15}; (2) facial motion is captured at close range, mostly frontal, and assuming little global head motion~\cite{Beeler:SIGGRAPH2010,ghosh2011multiview, Beeler:SIGGRAPH2011, bradley2010high, valgaerts2012lightweight}; (3) finger motion is also captured at very close distances from hands, where the hand regions are dominant in the sensor measurements~\cite{Oikonomidis-12, Tompson-14a, Sridha-15, Tzionas-16}. These configurations make it difficult to analyze these gestures in the context of social communication.

In this paper, we present a novel approach to capture the motion of the principal body parts for multiple interacting people (see Fig.~\ref{fig:teaser2}). The fundamental difficulty of such capture is caused by the scale differences of each part. For example, the torso and limbs are relatively large and necessitate coverage over a sufficiently large working volume, while fingers and faces, due to their smaller feature size, require close distance capture with high resolution and frontal imaging. With off-the-shelf cameras, the resolution for face and hand parts will be limited in a room-scale, multi-person capture setup. 

To overcome this sensing challenge, we use two general approaches: (1) we leverage keypoint detection (e.g., faces~\cite{Torre15}, bodies~\cite{Wei2016,cao2016realtime,Newell-16}, and hands~\cite{simon2017hand}) in multiple views to obtain 3D keypoints, which is robust to multiple people and object interactions; (2) to compensate for the limited sensor resolution, we present a novel generative body deformation model, which has the ability to express the motion of the each of the principal body parts. In particular, we describe a procedure to build an initial body model, named ``Frankenstein", by seamlessly consolidating available part template models~\cite{Loper2015,cao2014facewarehouse} into a single skeleton hierarchy. We optimize this initialization using a capture of 70 people, and learn a new deformation model, named ``Adam", capable of additionally capturing variations of hair and clothing, with a simplified parameterization. We present a method to capture the total body motion of multiple people with the 3D deformable model. Finally, we demonstrate the performance of our method on various sequences of social behavior and person-object interactions, where the combination of face, limb, and finger motion emerges naturally. 

\section{Related Work}
Motion capture systems performed by tracking retro-reflective markers~\cite{woltring1973new} are the most widely used motion capture technology due to their high accuracy. Markerless motion capture methods~\cite{Gavrila-96, deAguiar-2008, Gall-09, Stoll-11} have been explored over the past two decades to achieve the same goal without markers, but they tend to implicitly admit that their performance is inferior by treating the output of marker based methods as a ground truth or an upper bound. However, over the last few years, we have witnessed a great advance in key point detections from images (e.g., faces~\cite{Torre15}, bodies~\cite{Wei2016,cao2016realtime,Newell-16}, and hands~\cite{simon2017hand}), which can provide reliable anatomical landmark measurements for markerless motion capture methods~\cite{Elhayek-15,joo2016panoptic,simon2017hand}, while the performance of marker based methods relatively remains the same with their major disadvantages including: (1) a necessity of sparsity in marker density for reliable tracking which limits the spatial resolution of motion measurements, and (2) a limitation in automatically handling occluded markers which requires an expensive manual clean-up. Especially, capturing high-fidelity hand motion is still challenging in marker-based motion capture systems due to the severe self-occlusions of hands~\cite{zhao2012combining}, while occlusions are implicitly handled by guessing the occluded parts with uncertainty using the prior learnt from a large scale dataset~\cite{simon2017hand}. Our method shows that the markerless motion capture approach potentially begins to outperform the marker-based counterpart by leveraging the learning based image measurements. As an evidence we demonstrate the motion capture from total body, which has not been demonstrated by other existing marker based methods. In this section, we review the most relevant markerless motion capture approaches to our method.

Markerless motion capture largely focuses on the motion of the torso and limbs. The standard pipeline is based on a multiview camera setup and tracking with a 3D template model~\cite{Liu-2013, Gavrila-96, Cheung-05, Bregler-04, Kehl-06, Corazza-10, Vlasic-08, Brox-10, Stoll-11, deAguiar-2008, Furukawa-2008, Elhayek-15}. In this approach, motion capture is performed by aligning the 3D template model to the measurements, which distinguish the various approaches and may include color, texture, silhouettes, point clouds, and landmarks. A parallel track of related work therefore focuses on capturing and improving body models for tracking, for which a highly controlled multiview capture system---specialized for single person capture---is used to build precise models. With the introduction of commodity depth sensors, single-view depth-based body motion capture became a popular direction~\cite{Baak-13, Shotton2011}. A recent collection of approaches aims to reconstruct 3D skeletons directly from monocular images, either by fitting 2D keypoint detections with a prior on human pose~\cite{Zhou2015,Bogo2016} or getting even closer to direct regression methods~\cite{Zhou2016,Mehta2017,tome2017lifting}.

Facial scanning and performance capture has been greatly advanced over the last decade. There exist multiview based methods showing excellent performance on high-quality facial scanning~\cite{Beeler:SIGGRAPH2010,ghosh2011multiview} and facial motion capture~\cite{Beeler:SIGGRAPH2011, bradley2010high, valgaerts2012lightweight}. Recently, light-weighed systems based on a single camera show a compelling performance by leveraging morphable 3D face model on 2D measurements\cite{garrido-tog-2013, Torre15, li2013realtime, thies2016face2face, cao2014facewarehouse, cao2015real, wu2016anatomically}. Hand motion captures are mostly lead by single depth sensor based methods~\cite{Oikonomidis-12, Tang-14, Tompson-14a, Keskin-12,Xu-13,Sun-15,Wan-16, Sridhar-13, Sharp-15, Sridha-15, Tzionas-16, Ye-16}, with few exceptions based on multi-view systems~\cite{Ballan-12, Sridhar-13, MANO:SIGGRAPHASIA:2017}. In this work, we take the latter approach and use the method of~\cite{simon2017hand} who introduced a hand keypoint detector for RGB images which can be directly applicable in multiview systems to reconstruct 3D hand joints. 

As a way to reduce the parameter space and overcome the complexity of the problems, generative 3D template models have been proposed in each field, for example the methods of \cite{anguelov2005scape, Loper2015, pons2015dyna} in body motion capture, the method of \cite{cao2014facewarehouse} for facial motion capture, and very recently, the combined body+hands model of Romero et al.~\cite{MANO:SIGGRAPHASIA:2017}. A generative model with  expressive power for total body motion has not been introduced.

\begin{figure}[t]
	\includegraphics[width=\columnwidth]{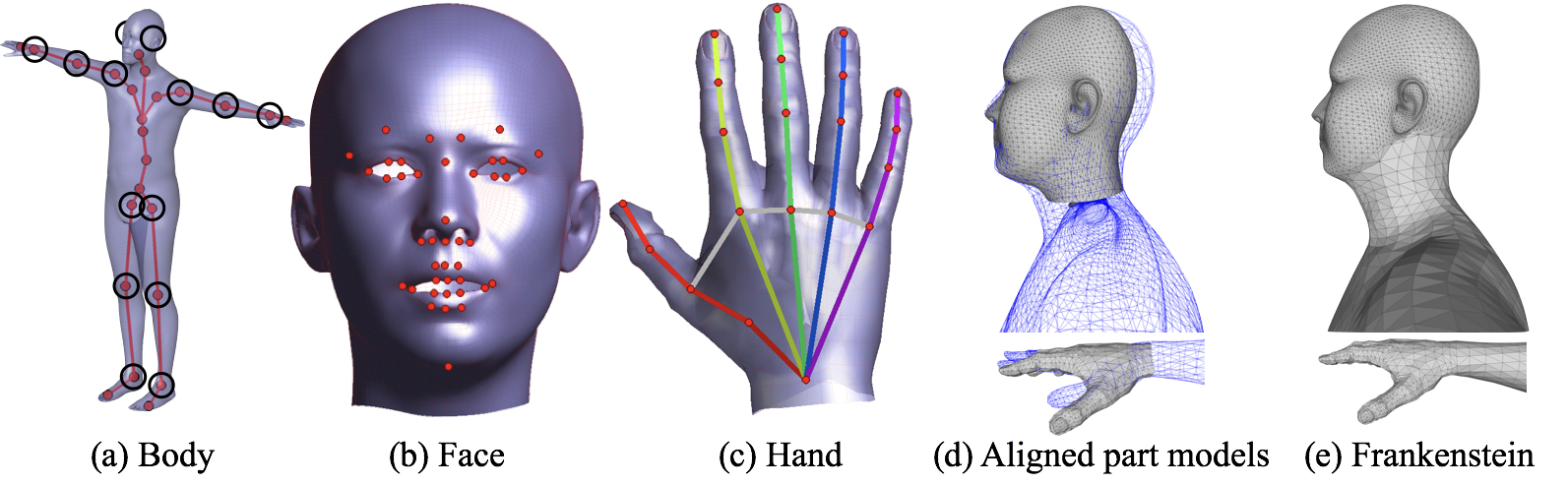}
	\caption{Part models and a unified Frankenstein model. (a) The body model~\cite{Loper2015}; (b) the face model~\cite{cao2014facewarehouse}; and (c) a hand rig, where red dots have corresponding 3D keypoints reconstructed from detectors in (a-c). (d) Aligned face and hand models (gray meshes) to the body model (the blue wireframe mesh); and (e) the seamless Frankenstein model.}
	\label{fig:frankenstein_part_aligned}
\end{figure}

\section{Frankenstein Model}

The motivation for building the Frankenstein body model is to leverage existing part models---SMPL~\cite{Loper2015} for the body, FaceWarehouse~\cite{cao2014facewarehouse} for the face, and an artist-defined hand rig---each of which capture shape and motion details at an appropriate scale for the corresponding part.
This choice is not driven merely by the free availability of the component models: note that due to the trade-off between image resolution and field of view of today's 3D scanning systems, scans used to build detailed face models will generally be captured using a different system than that used for the rest of the body.
For our model, we merge all transform bones into a single skeletal hierarchy but keep the native parameterization of each component part to express identity and motion variations, as explained below. As the final output, the Frankenstein model produces motion parameters capturing the total body motion of humans, and generates a seamless mesh by blending the vertices of the component meshes. 

\subsection{Stitching Part Models}

The Frankenstein model $M^U$ is parameterized by motion parameters $\boldsymbol{\theta}^U$, shape (or identity) parameters $\boldsymbol{\phi}^U$, and a global translation parameter $\mathbf{t}^U$,
\begin{align}
\mathbf{V}^U = M^U (\boldsymbol{\theta}^U, \boldsymbol{\phi}^U, \mathbf{t}^U ),
\end{align}
where $\mathbf{V}^U$ is a seamless mesh expressing the motion and shape of the target subject.

The motion and shape parameters of the model are a union of the part models' parameters:
\begin{align}
\boldsymbol{\theta}^U = \{ \boldsymbol{\theta}^B, \boldsymbol{\theta}^F, \boldsymbol{\theta}^{LH}, \boldsymbol{\theta}^{RH}  \}, \\
\boldsymbol{\phi}^U = \{ \boldsymbol{\phi}^B, \boldsymbol{\phi}^F, \boldsymbol{\phi}^{LH}, \boldsymbol{\phi}^{RH}  \},
\end{align}
where the superscripts represent each part model: $B$ for the body model, $F$ for the face model, $LH$ for for the left hand model, and $RH$ for the right hand model. Each of the component part models maps from a subset of the above parameters to a set of vertices, respectively, $\mathbf{V}^B \,{\in}\, \mathds{R}^{N^B{\times}3}$, $\mathbf{V}^F \,{\in}\, \mathds{R}^{N^F{\times}3}$, $\mathbf{V}^{LH} \,{\in}\, \mathds{R}^{N^H{\times}3}$, and $\mathbf{V}^{RH} \,{\in}\, \mathds{R}^{N^H{\times}3}$, where the number of vertices of each mesh part is $N^B{=}6890$, $N^H{=}2068$, and $N^F{=}11510$. The final mesh of the Frankenstein model, $\mathbf{V}^U {\in} \mathds{R}^{N^U{\times}3}$, is defined by linearly blending them with a matrix $\mathbf{C} \in \mathds{R}^{N^U\times(N^B{+}N^F{+}2N^H)}$: 
\begin{align}
\mathbf{V}^U = \mathbf{C} 
\left[
\begin{array}{c}
\left({\mathbf{V}^B}\right)^T
\left({\mathbf{V}^F}\right)^T
\left({\mathbf{V}^{LH}}\right)^T
\left({\mathbf{V}^{RH}}\right)^T
\end{array} 
\right]^T,
\end{align}
where $T$ denotes the transpose of a matrix. Note that $\mathbf{V}^U$ has fewer vertices than the sum of part models because there are redundant parts in the body model (e.g., face and hands of the body model). In particular, our final mesh has $N^U{=}18540$ vertices. Figure \ref{fig:frankenstein_part_aligned} shows the part models which are aligned by manually clicking corresponding points between parts, and also shows the final mesh topology of Frankenstein model at the mean shape in the rest pose. 
The blending matrix $\mathbf{C}$ is a very sparse matrix and most rows have a single column set to one with zeros elsewhere, simply copying the vertex locations from the corresponding part models with minimal interpolation at the seams.

%
In the Frankenstein model, all parts are rigidly linked by a single skeletal hierarchy. This unification is achieved by substituting the hands and face branches of the SMPL body skeleton with the corresponding skeletal hierarchies of the detailed part models. All parameters of the Frankenstein model are jointly optimized for motion tracking and identity fitting. The parameterization of each of the part models is detailed in the following sections.


\subsection{Body Model}

For the body, we use the SMPL model~\cite{Loper2015} with minor modifications. 
In this section, we summarize the salient aspects of the model in our notation. The body model, $M^B$, is defined as follows,
\begin{align}
\mathbf{V}^B = M^B (\boldsymbol{\theta}^B, \boldsymbol{\phi}^B, \boldsymbol{t}^B )
\end{align}
with $\mathbf{V}^B = \{ \mathbf{v}^B_i\}_{i=1}^{N^B}$. 
The model uses a template mesh of $N^B{=}6890$ vertices, where we denote the $i$-th vertex as $\mathbf{v}^B_i\in\mathds{R}^3$. 
The vertices of this template mesh are first displaced by a set of blendshapes describing the {\em identity} or body shape. Given the vertices in the rest pose, the posed mesh vertices are obtained by linear blend skinning using transformation matrices $\mathbf{T}^B_j\in \textrm{SE(3)}$ for each of $J$ joints,
\begin{align}
\mathbf{v}_i^B= \mathbf{I}_{3\times 4} \cdot \sum_{j=1}^{J^B} w^B_{i,j}\mathbf{T}^B_j\begin{pmatrix} \mathbf{v}^{B0}_i + \sum_{k=1}^{K_b} \mathbf{b}^k_{i} \phi^B_k \\ 1 \end{pmatrix},
\label{eq:full_lbs_pose}
\end{align}
where $\mathbf{b}^k_{i}\in\mathds{R}^3$ is the $i$-th vertex of the $k$-th blendshape, $\phi^B_k$ is the $k$-th shape coefficient in $\boldsymbol{\phi}^B\in\mathds{R}^{K_b}$ with $K_b{=}10$ the number of identity body shape coefficients, and $\mathbf{v}^{B0}_i$ is the $i$-th vertex of the mean shape. The transformation matrices $\mathbf{T}^B_j$ encode the transform for each joint $j$ from the rest pose to the posed mesh in world coordinates, which is constructed by following skeleton hierarchy from the root joint with pose parameter $\boldsymbol{\theta}^B$ (see~\cite{Loper2015}). The $j$-th pose parameter $\theta^B_j$ is the angle-axis representation of the relative rotation of joint $j$ with respect to its parent joints. $w^B_{i,j}$ is the weight with which transform $\mathbf{T}^B_j$ affects vertex $i$, with $\sum_{j=1}^Jw^B_{i,j}{=}1$ and $\mathbf{I}_{3\times 4}$ is the $3{\times} 4$ truncated identity matrix to transform from homogenous coordinates to a $3$ dimensional vector. We use $J^B{=}21$ with $\boldsymbol{\theta}^B \,{\in}\, \mathds{R}^{21{\times}3}$, ignoring the last joint of each hand of the original body model. For simplicity, we do not use the pose-dependent blendshapes.

\subsection{Face Model}
\label{subsection:face}
As a face model, we build a generative PCA model from the FaceWarehouse dataset~\cite{cao2014facewarehouse}. Specifically, the face part model, $M^F$, is defined as follows,
\begin{align}
\mathbf{V}^F = M^F (\boldsymbol{\theta}^F, \boldsymbol{\phi}^F, \mathbf{T}^F ),
\end{align}
with $\mathbf{V}^F = \{ \mathbf{v}^F_i\}_{i=1}^{N^F}$, where the $i$-th vertex is $\mathbf{v}^F_i\in\mathds{R}^3$, and $N^F{=}11510$. The vertices are represented by the linear combination of the subspaces:
\begin{align}
\hat{\mathbf{v}}_i^F = \mathbf{v}^{F0}_i + \sum_{k=1}^{K_{f}} \mathbf{f}^k_{i} \phi^F_k  + \sum_{s=1}^{K_{e}} \mathbf{e}^s_{i} \theta^F_s
\label{eq:face_shape}
\end{align}
where, as before, $\mathbf{v}^{F0}_i$ denotes $i$-th vertex of the mean shape, and $\phi^F_k$ and $\theta^F_s$ are $k$-th face shape identity (shape) and $s$-th facial expression (pose) parameters respectively. Here, $\mathbf{f}^k_i\in\mathds{R}^3$ is the $i$-th vertex of the $k$-th identity blendshape ($K_{f}=150$), and $\mathbf{e}^s_i\in\mathds{R}^3$ is the $i$-th vertex of the $s$-th expression blendshape ($K_{e}=200$). 

Finally, a transformation $\mathbf{T}^F$ brings the face vertices into world coordinates. To ensure that the face vertices transform in accordance to the rest of the body, we manually align the mean face $\mathbf{v}^{F0}_i$ with the body mean shape, as shown in Fig.~\ref{fig:frankenstein_part_aligned}. This way, we can apply the transformation of the body model's head joint $\mathbf{T}^B_{j=F}(\boldsymbol{\theta}^B)$ as a global transformation for the face model in Eq.~\ref{eq:face_pose}. However, to keep the face in alignment with the body, an additional transform matrix $\mathbf{\Gamma}^F \in \textrm{SE(3)}$ is required to compensate for displacements in the root location of the face joint due to body shape changes in Eq.~\ref{eq:full_lbs_pose}. 

Finally, each face vertex position is given by:
\begin{align}
\mathbf{v}^F_i = \mathbf{I}_{3\times 4} \cdot \mathbf{T}^B_{j=F} \cdot \mathbf{\Gamma}^F \begin{pmatrix} \hat{\mathbf{v}}^F_i \\ 1 \end{pmatrix},
\label{eq:face_pose}
\end{align}
where the transform $\mathbf{\Gamma}^F$, directly determined by the body shape parameters $\boldsymbol{\phi}^B$, aligns the face model with the body model.

\subsection{Hand Model}
We use an artist rigged hand mesh. Our hand model has $J^H{=}16$ joints and the mesh is deformed via linear blend skinning. The hand model has a fixed shape, but we introduce scaling parameters for each bone to allow for different finger sizes.
The transform for each joint $j$ is parameterized by the Euler angle rotation
with respect to its parent, $\boldsymbol{\theta}_j\in\mathds{R}^3$, and an additional anisotropic scaling factor along each axis, $\boldsymbol{\phi}_j\in\mathds{R}^3$. Specifically, the linear transform for joint $j$ in the bone's local reference frame becomes $\operatorname{eul}(\boldsymbol{\theta}_j)\cdot \operatorname{diag}(\boldsymbol{s}_j)$, 
where $\operatorname{eul}(\boldsymbol{\theta}_j)$ converts from an Euler angle representation to a $3\times 3$ rotation matrix and $\operatorname{diag}(\boldsymbol{\phi}_j)$ is the $3\times 3$ diagonal matrix with the $X$,$Y$,$Z$ scaling factors $\mathbf{\phi}_j$ on the diagonal. The vertices of the hand in world coordinates are given by LBS with weights $w^H_{i,j}$: 
\begin{align}
\mathbf{v}_i= \mathbf{I}_{3\times 4} \cdot  \mathbf{T}^B_{j=H} \cdot \mathbf{\Gamma}^H \cdot \sum_{j=1}^J w^H_{i,j}\mathbf{T}^H_j \begin{pmatrix} \mathbf{v}_i^0 \\ 1 \end{pmatrix}.
\label{eq:lbs_hand}
\end{align}
where $\mathbf{T}^H_j$ is each bone's composed transform (with all parents in the hierarchy), $\mathbf{T}^B_{j=H}$ is the transformation of the corresponding hand joint in the body model, and $\mathbf{\Gamma}^H$ is the transformation that aligns the hand model to the body model. As with the face, this transform depends on the shape parameters of the body model.



\section{Motion Capture with Frankenstein Model}

We fit the Frankenstein model to data to capture the total body motion, including the major limbs, the face, and fingers. Our motion capture method relies heavily on fitting mesh correspondences to 3D keypoints, which are obtained by triangulation of 2D keypoint detections across multiple camera views. To capture shape information we also use point clouds generated by multiview stereo reconstructions. Model fitting is performed by an optimization framework to minimize distances between corresponded model joints and surface points and 3D keypoint detections, and iterative closest point (ICP) to the 3D point cloud.

\subsection{3D Measurements}
\label{subsection:landmark_reconstruction}
\begin{figure}[t]	
	\includegraphics[width=\columnwidth]{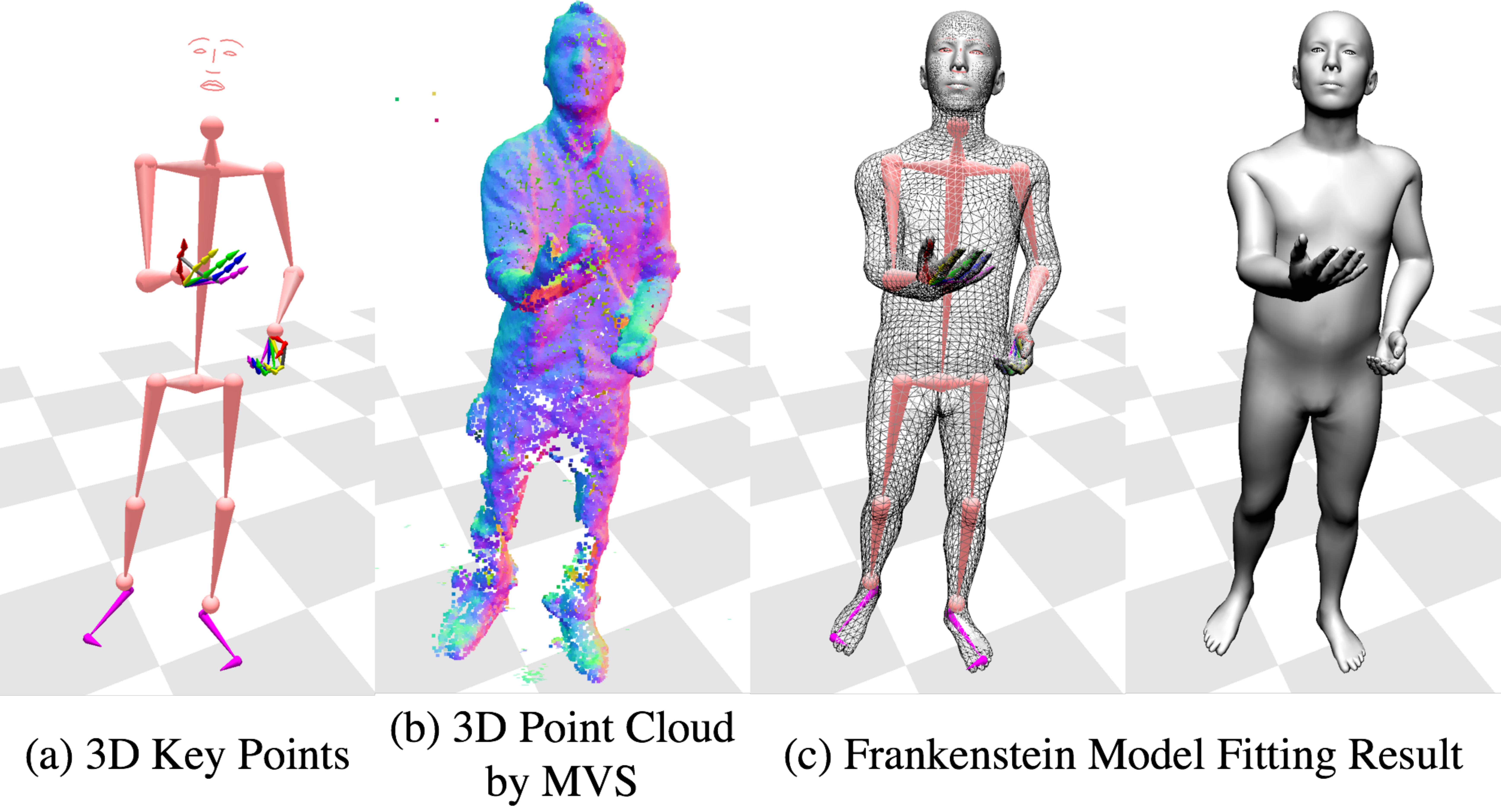}
	\caption{3D measurements and Frankenstein fitting result.}
	\label{fig:3dlandmarks_model_fitting}
\end{figure}

We incorporate two types of measurements in our framework as shown in Fig.~\ref{fig:3dlandmarks_model_fitting}: (1) corresponded 3D keypoints, which map to known joints or surface points on the mesh models (see Fig.~\ref{fig:frankenstein_part_aligned}), and (2) uncorresponded 3D points from multiview stereo reconstruction, which we match using ICP. 

\textbf{3D Body, Face, and Hand Keypoints:}  We use the OpenPose detector~\cite{openpose} in each available view, which produces 2D keypoints on the body with the method of~\cite{cao2016realtime}, and hand and face keypoints using the method of \cite{simon2017hand}. 3D body skeletons are obtained from the 2D detections using the method of~\cite{joo2016panoptic}, which uses known camera calibration parameters for reconstruction. The 3D hand keypoints are obtained by triangulating 2D hand pose detections, following the method of \cite{simon2017hand}, and similarly for the facial keypoints. Note that subsets of 3D keypoints can be entirely missing if there aren't enough 2D detections for triangulation, which can happen in challenging scenes with inter-occlusions or motion blur. 

\textbf{3D Feet Keypoints:} An important cue missing from the OpenPose detector are landmarks on the feet. For motion capture, this is an essential feature to prevent footskate, as well as to accurately determine the orientation of the feet. We therefore train a keypoint detector for the tip of the big toe, the tip of the little toe, and the ball of the foot. We annotate these 3 keypoints per foot in each of around 5000 person instances of the COCO dataset, and use the neural network architecture presented by ~\cite{Wei2016} with a bounding box around the feet determined by the 3D body detections\footnote{More details provided in the supplementary material.}.


%
%
%

\textbf{3D Point Clouds:} We use the commercial software Capturing Reality to obtain 3D point clouds from the multiview images, with associated point normals.

	

\subsection{Objective Function}

We initially fit every frame in the sequence independently. For clarity, we drop the time index from the notation and describe the process for a single frame, which optimizes the following cost function:
\begin{align}
\label{eq:fitting_franken}
E\big( \mathbf{\theta}^U, \mathbf{\phi}^U, \mathbf{t}^U \big) = E_\textrm{keypoints} + E_\textrm{icp} + E_\textrm{seam} +E_\textrm{prior}
\end{align}
\textbf{Anatomical Keypoint Cost:} the term $E_\textrm{keypoints}$ matches 3D keypoint detections which are in direct corresponce to our mesh models. This includes joints (or end effects) in the body and hands, and also contains points corresponding to the surface of the mesh (e.g., facial keypoints and the tips of fingers and toes). Both of these types of correspondence are expressed as combinations of vertices via a regression matrix $\mathbf{J}\,{\in}\,\mathds{R}^{C\times N^U}$, where $C$ denotes the number of correspondences and $N^U$ is the number of vertices in the model. Let $\mathcal{D}$ denote the set of available detections in a particular frame. The cost is then:
\begin{align}
E_\textrm{keypoints} = \lambda_\textrm{keypoints} \sum_{i\in\mathcal{D}}|| \mathbf{J}_i \mathbf{V} - \mathbf{y}_i^T ||^2,
\label{eq:detection_eq}
\end{align}
where $\mathbf{J}_i$ indexes a row in the correspondence regression matrix and represents an interpolated position using a small number of vertices, and $\mathbf{y}_i\,{\in}\,\mathds{R}^{3 \times 1}$ is the 3D detection. The $\lambda_\textrm{keypoints}$ is a relative weight for this term.

\textbf{ICP Cost:} The 3D point cloud measurements are not a priori in correspondence with the model meshes. We therefore establish their correspondence to the mesh using Iterative Closest Point (ICP) during each solver iteration. 
We find the closest 3D point in the point cloud to each of the mesh vertices,
\begin{align}
i^*= \operatorname{arg} \operatorname{min}_i  || \mathbf{x}_i - \mathbf{v}_j ||^2,
\end{align}
where $\mathbf{x}_{i^*}$ is the closest 3D point to vertex $j$, where $\mathbf{v}_j$ is a vertex\footnote{We do not consider some parts (around hands and face), as depth sensor resolution is too low to improve the estimate. These parts are defined as a mask.} in $\mathbf{V}^U$ of the Frankenstein model. To ensure that this is a correct correspondence, we use thresholds for the distance and normals during the correspondence search.

Finally, for each vertex~$j$ we compute the point-to-plane residual, i.e., the distance along the normal direction,
\begin{align}
E_\textrm{icp} = \lambda_\textrm{icp} \sum_{\mathbf{v}_j \in \mathbf{V}^U_t} \mathbf{n}(\mathbf{x}_{i^*})^T (\mathbf{x}_{i^*} - \mathbf{v}_j ),
\end{align}
where $\mathbf{n}(\cdot)\in\mathds{R}^3$ represents the point's normal, and $\lambda_\textrm{icp}$ is a relative weight for this term. 

\textbf{Seam Constraints:} The part models composing the Frankenstein model are rigidly linked by the skeletal hierarchy. However, the independent surface parameterizations of each of the part models may introduce discontinuities at the boundary between parts (e.g., a fat arm with a thin wrist). To avoid this artifact, we encourage the vertices around the seam parts to be close by penalizing differences between the last two rings of vertices around the seam of each part, and the corresponding closest point in the body model in the rest pose expressed as barycentric coordinates (see the supplementary materials for details).
%

\textbf{Prior Cost:} 
Depending on the number of measurements available in a particular frame, the set of parameters of $M^u$ may not be determined uniquely (e.g., the width of the fingers). More importantly, the 3D point clouds are noisy and cannot be well explained by the model due to hair and clothing, which are not captured by the SMPL and FaceWarehouse meshes, which can result in erroneous correspondences during ICP. Additionally, the joint locations of the models are not necessarily consistent with the annotation criteria used to train the 2D detectors. We are therefore forced to set priors over model parameters to avoid the model from overfitting to these sources of noise, $E_\textrm{prior} = E^F_\textrm{prior} + E^B_\textrm{prior} + E^H_\textrm{prior}$.
The prior for each part is defined by corresponding shape and pose priors, for which we use 0-mean standard normal priors for each parameter except for scaling factors, which are encouraged to be close to 1. Details and relative weights can be found in supplementary materials.




\subsection{Optimization Procedure}
The complete model is highly nonlinear, and due to the limited degrees of freedom of the skeletal joints, the optimization can get stuck in bad local minima. 
Therefore, instead of optimizing the complete model initially, we fit the model in phases, starting with a subset of measurements and strong priors that are relaxed as optimization progresses.

Model fitting is performed on each frame independently. To initialize the overall translation and rotation, we use four keypoints on the torso (left and right shoulders and hips) without using the ICP term, and with strong weight on the priors. Once the torso parts are approximately aligned, we use all available keypoints of all body parts, with small weight for the priors. The results at this stage already provide reasonable motion capture but do not accurately capture the shape (i.e., silhouette) of the subject. Finally, the entire optimization is performed including the ICP term to find correspondences with the 3D point cloud. We run the final optimization two times, finding new correspondences each time. For the optimization we use Levenberg-Marquardt with the Ceres Solver library~\cite{ceres-solver}.

\begin{figure}[t]    
\centering
	\includegraphics[width=0.9\columnwidth]{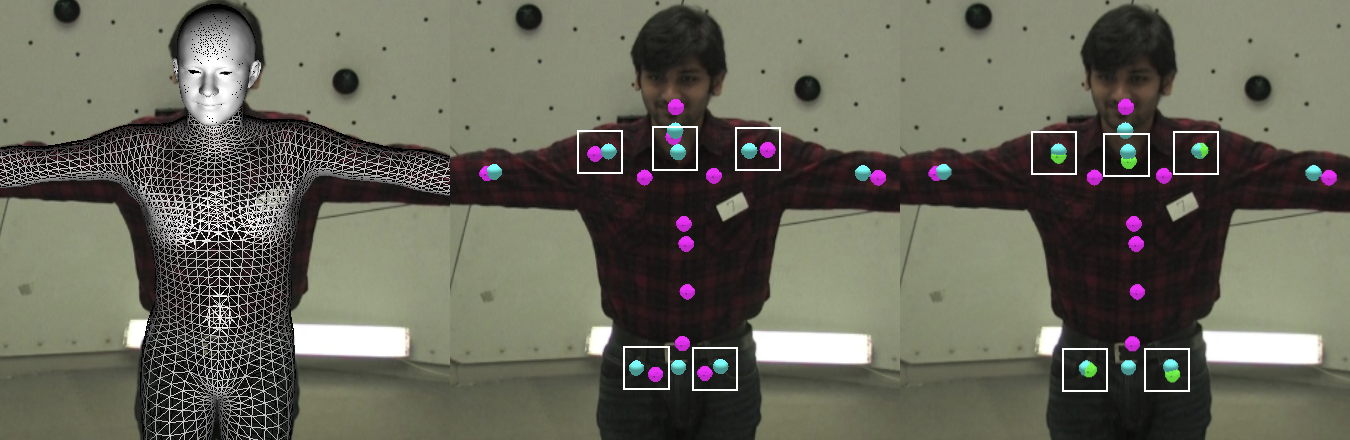}    
	\caption{Regressing detection target positions. (Left) The template model is aligned with target object. (Mid.) The torso joints of the template model (magenta) have discrepancy from the joint definitions of 3D keypoint detection (cyan). (Right) The newly regressed target locations (green) are more consistent with 3D keypoint detections.}
	\label{fig:jointRegression}
\end{figure}
%
%

\section{Creating Adam}
We derive a new model, which we call Adam, enabling total body motion capture with a simpler parameterization than the part-based Frankenstein model. In particular, this new model has a single joint hierarchy and a common parameterization for all shape degrees of freedom, tying together the face, hand, and body shapes and avoiding the need for seam constraints. To build the model, it is necessary to reconstruct the shape and the motion of all body parts (face, body, and hands) from diverse subjects where model can learn the variations. To do this, we leverage our Frankenstein model and apply it on a dataset of 70 subjects where each of them performs a short range of motion in a multiview camera system. We selected 5 frames for each person in different poses and use the the reconstruction results to build Adam. From the data, both joint location information and linear shape blendshapes are learnt. Because we derive the model from clothed people, the blendshapes explain some variations of them.


\subsection{Regressing Detection Targets}

There exists a discrepancy between the joint locations of the body model (e.g., SMPL model in our case) and the location of the keypoint detections (i.e., a model joint vs. a detection joint), as shown in Fig.~\ref{fig:jointRegression}. This affects mainly the shoulder and hip joints, which are hard to precisely annotate. This difference has the effect of pulling the Frankenstein model towards a bad fit even while achieving a low keypoint cost, $E_{\textrm{keypoints}}$.
We alleviate this problem by computing the relative location of the 3D detections with respect to the fitted mesh vertices by leveraging the the reconstructed 70 people data. This allows us to define new targets for the keypoint detection cost that, on average, are a better match for the location of the 3D detections with respect to the mesh model, as shown in Fig.~\ref{fig:jointRegression}. In particular, given the fitting results of 70 identities, we approximate the target 3D keypoint locations as a function of the final fitted mesh vertices following the procedure of~\cite{Loper2015} to find a sparse, linear combination of vertices that approximates the position of the target 3D keypoint. Note that we do not change the joint location used in the skeleton hierarchy during LBS deformation, only the regression matrices $\mathbf{J}_i$ in Eq.~\eqref{eq:detection_eq}. 
\subsection{Fitting Clothes and Hair}
The SMPL model captures the shape variability of human bodies, but does not account for clothing or hair. Similarly, the FaceWarehouse template mesh was not design to model hair. However, for the natural interactions that we are most interested in capturing, people wear everyday clothing and sport typical hairstyles. 
To learn a new set of linear blendshapes that better capture the rough geometry of clothed people and jointly model face, it is required to reconstruct the accurate geometry of the source data. For this purpose, we reconstruct the out-of-shape spaces in the reconstructed 70 people results by Frankenstein model fitting. For each vertex in the Frankenstein model, we write
\begin{align}
\tilde{\mathbf{v}}_i  =  \mathbf{v}_i  + \mathbf{n}(\mathbf{v}_i) \delta_i,
\end{align}
where $\delta_i \in \mathds{R} $ is a scalar displacement meant to compensate for the discrepancy between the Frankenstein model vertices and the 3D point cloud, along the normal direction at each vertex. 
We pose the problem as a linear system,
%
\begin{align}
\label{eq:delta_recon}
\begin{pmatrix} \mathbf{N}^T  \\ (\mathbf{W} \mathbf{L} \mathbf{N})^T  \end{pmatrix} \Delta
 = \begin{pmatrix} (\mathbf{P} - \mathbf{V}^U)^T \\ \mathbf{0} \end{pmatrix},
\end{align}
where $\Delta\in\mathds{R}^{N^U}$ contains the stacked per-vertex displacements, $\mathbf{V}^U$ are the vertices in the Frankenstein model, $\mathbf{P}\in\mathds{R}^{N^U\times 3}$ are corresponding point cloud points, $\mathbf{N}\in\mathds{R}^{N^U\times 3}$ contains the mesh vertex normals, and $\mathbf{L}\in\mathds{R}^{N^U\times N^U}$ is the Laplace-Beltrami operator to regularize the deformation. We also use a weight matrix $\mathbf{W}$ to avoid large deformations where the 3D point cloud has lower resolution than the original mesh, like details in the face and hands. 

\subsection{Building the Shape Deformation Space}

After $\Delta$ fitting, we warp each frame's surface to the rest pose, applying the inverse of the LBS transform. With the fitted surfaces warped to this canonical pose, we do PCA analysis to build a joint linear shape space that captures shape variations across the entire body. As in Section~\ref{subsection:face}, we separate the expression basis for the face and retain the expression basis from the FaceWarehouse model, as our MVS point clouds are of too low resolution to fit facial expressions.

This model now can have shape variation for all parts, including body, hand, and face. The model also includes deformation of hair and clothing. That is this model can substitute parameters of $\phi^F$, $\phi^B$, and $\phi^H$. 
\begin{align}
M^T (\mathbf{\theta}^T, \mathbf{\phi}^T, \mathbf{t}^T ) = \mathbf{V}^T
\end{align}
with $\mathbf{V}^T = \{ \mathbf{v}^T_i\}_{i=1}^{N^T}$ and $N^T{=}18540$. As in SMPL, the vertices of this template mesh are first displaced by a set of blendshapes in the rest pose, $\hat{\mathbf{v}}^T_i = \mathbf{v}^{T0}_i + \sum_{k=1}^{K_T} \mathbf{s}^k_{i} \phi^B_k,$ 
where $\mathbf{s}^k_{i}\in\mathds{R}^3$ is the $i$-th vertex of the $k$-th blendshape, $\phi^T_k$ is the $k$-th shape coefficients of $\mathbf{\phi^T}\in\mathds{R}^{K_b}$, and $K_T=40$ is the number of identity coefficients, $\mathbf{v}^{T0}$ is the mean shape and $\mathbf{v}^{T0}$ is its $i$-th vertex. However, these blendshapes now capture variation across the face, hands, and body. These are then posed using LBS as in Eq.~\eqref{eq:full_lbs_pose}. We define the joints and weights for LBS followoing the part models, which is further explained in the supplementary material. 

%
	
\subsection{Tracking with Adam}

The cost function to capture total body motion using Adam model is similar to Eqn.~\ref{eq:fitting_franken} without the seam term:
\begin{align}
\label{eq:fitting_adam}
E\big( \mathbf{\theta}^T, \mathbf{\phi}^T, \mathbf{t}^T\big) = E_\textrm{keypoints} + E_\textrm{icp} +E_\textrm{prior}.
\end{align}
However, Adam is much easier to use than Frankenstein, because  it only has a single type of shapes and pose parameters for all parts. Conceptually, it is based on the SMPL model parameterization, but with additional joints for the hands and facial expression blendshapes.

\textbf{Optical Flow Propagation}: While fitting each frame independently has benefits----it does not suffer from error accumulation and frames can be fit in parallel---it typically produces jittery motion. To reduce this jitter, we use optical flow to propagate the initial, per-frame fit to neighboring frames to find a smoother solution. More concretely, given the fitting results at the frame $t$, we propagate this mesh to  frames $t{-}1$ and $t{+}1$ using optical flow at each vertex, which is triangulated into 3D using the method of~\cite{Joo2014}. Therefore, each vertex has at most three candidate positions: the original mesh, and the forward and backward propagated vertices (subject to a forward-backward consistency check). Given these propagated meshes, we reoptimize the model parameters by using all propagated mesh vertices as additional keypoints to find a compromise mesh. 
We run this process multiple times (3, in our case), to further reduce jitter and fill in frames with missing detections.


\begin{figure}[t]	
    \centering
	\includegraphics[width=0.9\columnwidth]{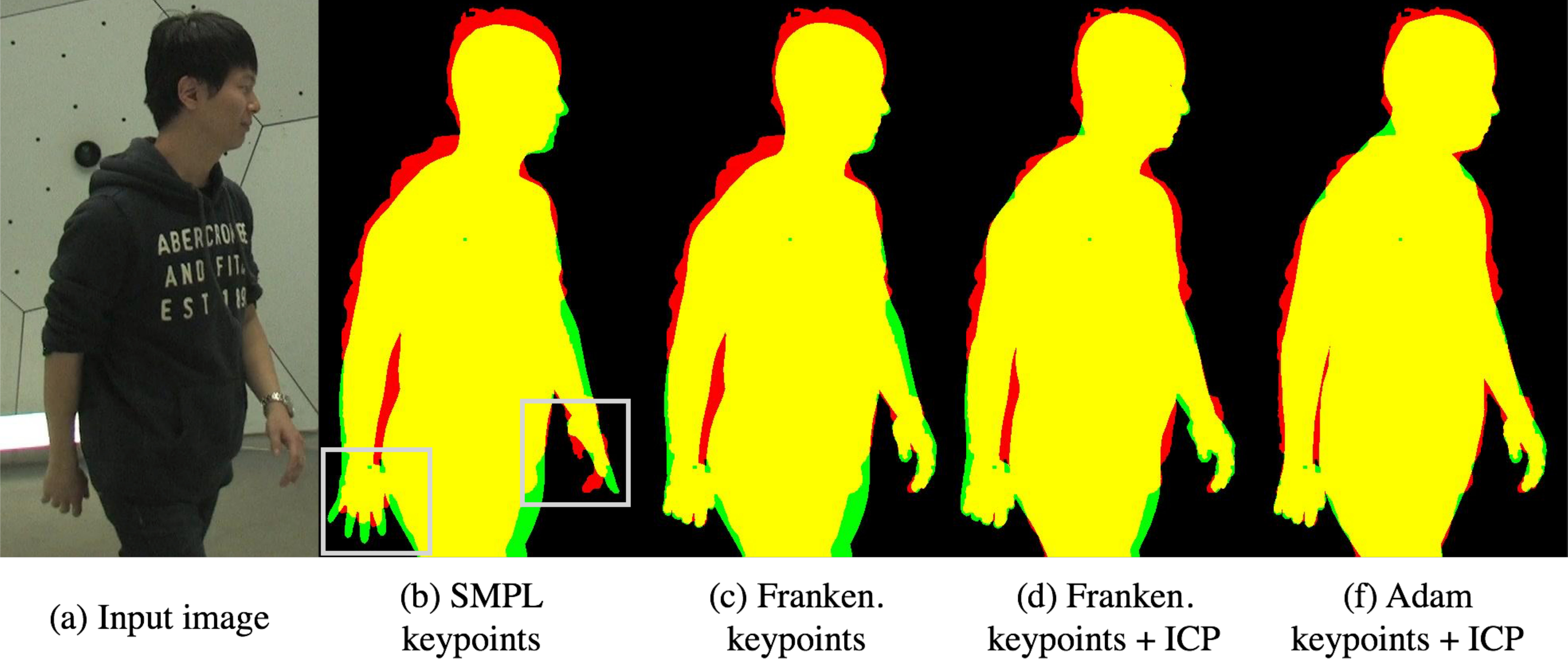}
	\includegraphics[width=0.9\columnwidth]{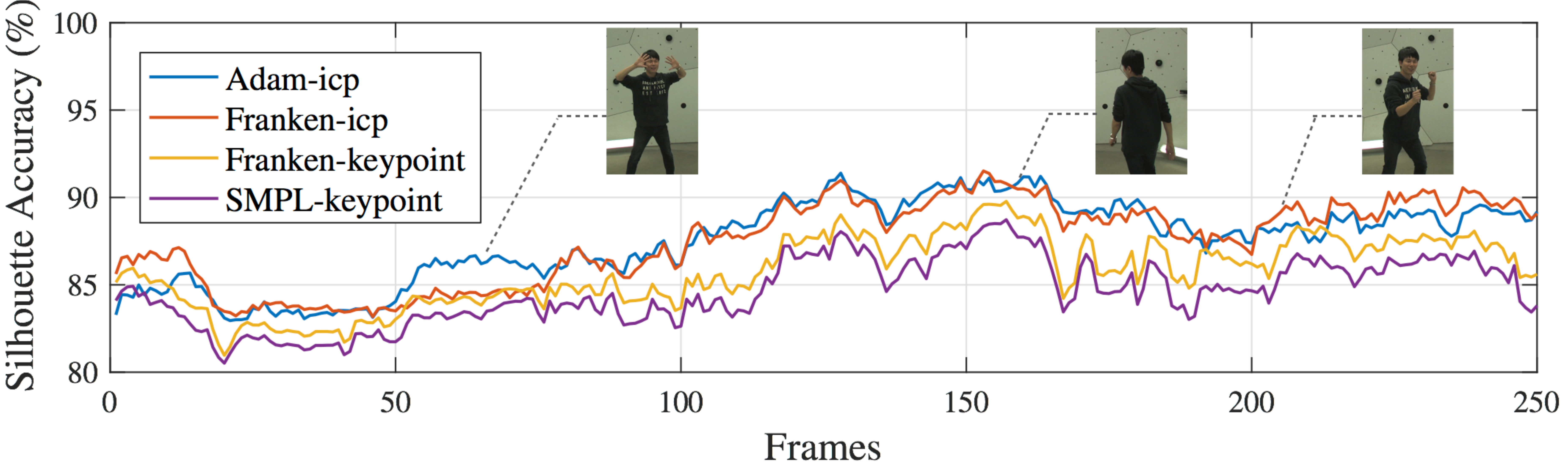}
	\caption{(Top) Visualization of silhouette from different methods with Ground-truth. The ground truth is drawn on red channel and the rendered silhouette masks from each model is drawn on green channel. Thus, the correctly overlapped region is shown as yellow color.; (Bottom) Silhouette accuracy compared to the ground truth silhouette.}
	\label{fig:quant_silhoutte_vis}
\end{figure}

\begin{table} [t]	
\centering
\scriptsize
	\caption{Accuracy of Silhouettes from different models}\label{Table:quant_silhoette}
	\begin{tabular}{c|c|c|c|c}
		\hline 
		& {SMPL\cite{Loper2015}} & {Franken} & {Franken ICP} & {Adam ICP} \tabularnewline
		\hline 
		Mean &  84.79\% & 85.91\% & 87.68\% & 87.74\%  \tabularnewline
		\hline 
		Std.  &  4.55  & 4.57 & 4.53  & 4.18  \tabularnewline
		\hline 
	\end{tabular} 
	\label{table:quant_sillouette}
\end{table}


\begin{figure*}[t]
	\includegraphics[width=\textwidth]{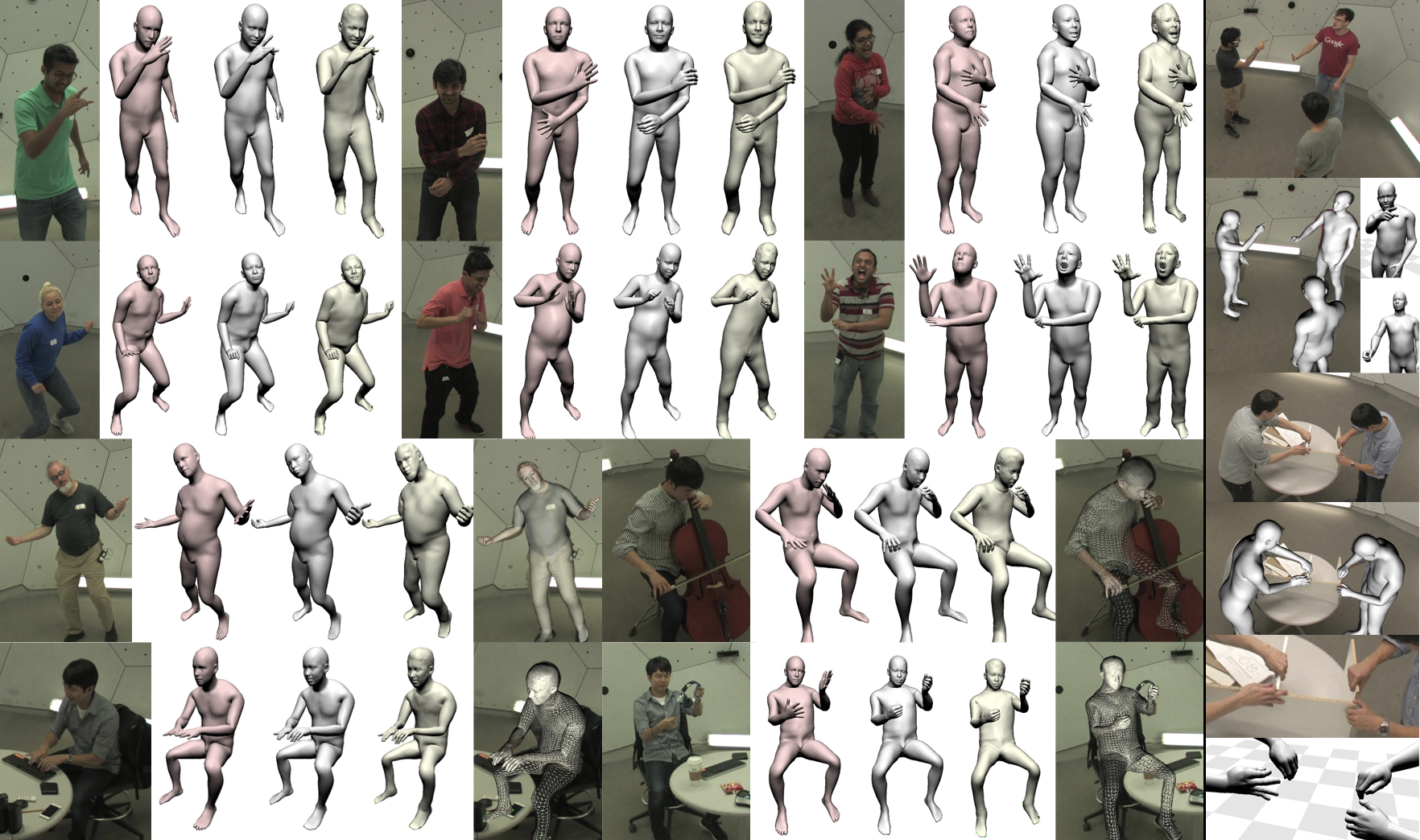}
	\caption{Total body reconstruction results on various human body motions. For each example scene, the fitting results from three different models are shown by different colors (pink for SMPL~\cite{Loper2015}, silver for Frankenstein, and gold for Adam). 
	}
	\label{fig:qualitativeResults}
\end{figure*}

\section{Results}
We perform total motion capture using our two models, Frankenstein and Adam, on various challenging sequences. For experiments, we use the dataset captured in the CMU Panoptic Studio~\cite{Joo-15}. We use 140 VGA cameras to reconstruct 3D body keypoints, 480 VGA cameras for feet, and 31 HD cameras for faces and hands keypoints, and 3D point clouds. We compare the fits produced by our models with the body-only SMPL model~\cite{Loper2015}. 

\subsection{Quantitative Evaluation}
We evaluate how well each model can match a moving person by measuring overlap with the ground truth silhouette across 5 different viewpoints for a 10 second range of motion sequence. To obtain the ground truth silhouette, we run a background subtraction algorithm using a Gaussian model for the background of each pixel with a post-processing to remove noise by morphological transforms. As an evaluation metric, we compute the percentage of overlapping region compared to the union between the GT silhouettes and the rendered forground masks after fitting each model. Here, we compare the fitting results of 3 different models: SMPL, our Frankenstein, and our Adam models. An example result is shown in Figure~\ref{fig:quant_silhoutte_vis}, and the results are shown in Fig~\ref{fig:quant_silhoutte_vis} and Table~\ref{table:quant_sillouette}. We first compare accuracy between SMPL and Frankenstein model by using only 3D keypoints as measurement cues. The major source of improvement of Frankenstein over SMPL is in the articulated hand model (by construction, the body is almost identical), as seen in Fig.~\ref{fig:quant_silhoutte_vis} (a). Including ICP term as cues provides better accuracy. Finally in the comparison between our two models, they show almost similar performance. Ideally we expect the Adam outperforms Frankenstein because it has more expressive power for hair and clothing, and it shows it shows better performance in a certain body shape (frame 50-75 in Fig~\ref{fig:quant_silhoutte_vis}). However, Adam sometimes produces artifacts showing lower accuracy; it tends to generate thinner legs, mainly due to poor 3D point cloud reconstructions on the source data on which Adam is trained. However, Adam is simpler for total body motion capture purpose and has potential to be improved once a large scale dataset is available with more optimized capture setup.




\subsection{Qualitative Results}
We run our method on sequences where face and hand motions are naturally emerging with body motions. The sequences include short range of motions for 70 people used to build Adam, social communications of multiple people, a furniture building sequence with dexterous hand motions, musical performances such as cello and guitars, and commonly observable daily motions such as keyboard typing. Most of these sequences are rarely demonstrated in previous markerless motion capture methods since capturing subtle details are the key to achieve the goal.  The example results are shown in Figure~\ref{fig:qualitativeResults}. Here, we also qualitatively compare our models (in silver color for Frankenstein, and gold for Adam) with the SMPL model (in pink)~\cite{Loper2015}. It should be noted that the total body motion capture results based on our models produce much better realism for the scene by capturing the subtle details from hands and faces. Our results are best shown in the accompanying videos.

\section{Discussion}
We present the first markerless method to capture total body motion including facial expression, coarse body motion from torso and limbs, and hand gestures at a distance. To achieve this, we present two types of models which can express motion in each of the parts. Our reconstruction results show compelling and realistic results, even when using only sparse 3D keypoint detections to drive the models.

{\small
\bibliographystyle{ieee}
\bibliography{tbc-bibliography}
}

\end{document}